\documentclass[a4paper]{article}
\pdfoutput=1
\usepackage{amsmath}
\usepackage{amsfonts}
\usepackage{amssymb}
\usepackage{algorithm}
\usepackage{algorithmic}
\usepackage[latin1]{inputenc}
\usepackage{xspace}
\usepackage[pdftex]{graphicx}
\usepackage{graphicx,caption}
\usepackage{url}
\usepackage{subfig}
\usepackage{booktabs}
\usepackage{url}

\usepackage{listings}
\usepackage{xcolor}
\definecolor{Gray}{gray}{0.70}
\usepackage[
	pdfauthor={Videh Seksaria},
	pdftitle={Multimodal Optimization by Sparkling Squid Populations},
	pdfsubject={Technical Report},
	pdfkeywords={},
	pdfcreator={pdflatex},
	colorlinks=false,
	linkcolor={black},
	citebordercolor={white},
	linkbordercolor={white},
	urlbordercolor={white},
        bookmarksopen={true}
]{hyperref}
\usepackage{nccfoots}
\usepackage{url}

\begin{document}
\begin{center}
\textbf{\Large  Multimodal Optimization by Sparkling Squid Populations}\\[10pt]
  Videh Seksaria \\
  Lexington High School \\
  Department of Computer Science  \\
  01vssv10@gmail.com
\end{center}
\begin{abstract}
The swarm intelligence of animals is a natural paradigm to apply to optimization problems. Ant colony, bee colony, firefly and bat algorithms are amongst those that have been demonstrated to efficiently to optimize complex constraints. This paper proposes the new Sparkling Squid Algorithm (SSA) for multimodal optimization, inspired by the intelligent swarm behavior of its namesake. After an introduction, formulation and discussion of its implementation, it will be compared to other popular metaheuristics. Finally, applications to well - known problems such as image registration and the traveling salesperson problem will be discussed.
\end{abstract}
\tableofcontents
\pagebreak
\section{Introduction}
\label{sec:intro}
Optimization problems are formulated based on the practical situations in which they occur in. The resultant model is grossly unfit for processing by classical algorithms in many ways. Primarily, there are few simplifying assumptions that can be made. Such algorithms often require the constraints to be eased, which can impact the quality of the solution produced. In addition, there efficiency depends on the complexity of the objective and scope and structure of the solution space. Instead of providing a general solution, they offer a specific solution that can be rendered useless by adding a single variable \cite{3,12,22}. This makes classical algorithms a poor choice for large, non - linear optimization tasks \cite{14}.

A more fit design paradigm is that of the multipurpose heuristic (metaheuristics) algorithm. Crafted to be able to solve an entire class of problems, metaheuristics are fast, and flexible to be tailored to precise specifications. These algorithms use iterative refinement and evolutionary methods to improve a solution candidate with regard to a fitness function \cite{28,21}. One particular class of interest is based on swarm intelligence. Swarm intelligence uses the cooperative behavior of animal societies to design algorithms. As in the real world, this allows a large solution space to be quickly searched, while also providing fault tolerance and parallelism \cite{15,2,8,7,10}.
	
Though diverse in their inspiration throughout nature, swarm algorithms depend on individual orientation and a fair division of labor. The ability for individuals to react to external stimuli also removes the need of processing cycle - consuming central hive - mind. Popular examples of such algorithms include particle swarm (PSO), ant colony, artificial bee colony (ABC), firefly (FA), and various cellular networks.  These methods have been shown to be superior to previous optimization strategies such as genetic algorithms (GA) and simulated annealing (SA). Other swarm strategies based on physical nature (as opposed to biological nature) include water droplets, electromagnetism, river dynamics, and musical based metrics \cite{12,22,26,53,10,24,13,32,19,1,33,9,5}. In addition to being more accurate, these strategies tend to be easier to implement and more robust. More recent advancements like particle swarm algorithms also have the added benefit of only using real numbers in computation as opposed to complex mutation functions as in genetic algorithms \cite{27,25,6}.
	
In this work, the sparkling squid algorithm (SSA) will be introduced and its performance investigated when compared with other relevant algorithms. Before outlining the sparkling squid algorithm, the particle swarm algorithm will be briefly outlined.

The rest of the paper is organized as follows. In section 2, the particle swarm algorithm is described, and various applications are introduced. In sections 3 and 4, the sparkling squid algorithm is defined and its behavior studied. In section 5, it is compared with pertinent algorithms, and evaluated in section 6.
\section{Optimization}
\label{sec:optim}
Particle swarm optimization is an optimization paradigm which travels the search space while attempting to optimize an objective function. It does so by tweaking the motion of individual particles (agents) as they traverse the space. The idea originates from the study of the swarming behavior of fish or birds. Though many variants of PSO exist, the idea by Eberhart and Kennedy (1995) is discussed below.

As the name suggest, PSO is a strategy for finding the values of parameters that give the optimum value of an objective function. This is critical if one wishes to maximize a certain value, such as profit or yield. That is:
\begin{center}    
\begin{equation}
\textrm{Given }f:{{\mathbb R}}^n\to {\mathbb R}
\end{equation}
\begin{equation}
\textrm{Find }\widehat{{\mathbf x}}\in {{\mathbb R}}^n,\ f\left(\widehat{{\mathbf x}}\right)\le f\left({\mathbf x}\right),\ {\mathbf x}\in {{\mathbb R}}^{{\mathbf n}}
\end{equation}
\end{center}
Of course, is one wishes to minimize ($\widehat{{\mathbf x}}\in {{\mathbb R}}^n,\ f\left(\widehat{{\mathbf x}}\right)\ge f\left({\mathbf x}\right),\ {\mathbf x}\in {{\mathbb R}}^{{\mathbf n}}{\mathbf )}$, they can maximize the value of $-f$. The domain ${{\mathbb R}}^n$of $f$z represents the search (parameter) space. Every element is a possible o, and so is referred to as a candidate solution. The dimension of this space is equivalent to the number of variables, EQ, while the fitness space of the fitness function has only a single dimension\cite{8}. 

This problem is trivial to solve if we are sure of the function $f$. However, in optimization tasks, the objective function within a ``black - box'', which prevents the methods of calculus from being applicable. One also may have to content with constrained optimization tasks where additional constraints such as positivity or primality are enforced\cite{14}. 

The simplest example of an optimization task can be seen below in figure 1. The relevant interval of a function f is shown. This function maps the candidate solutions on the x - axis, to the result of the objective function on the y - axis. In general, this fitness landscape shows the one - dimensional parameter space versus the one - dimensional fitness value. In addition, there is the presence of a local maximum, the maximum if we restrict the domain of candidate solutions, and a separate global maximum. Many variants of the PSO algorithm exist to find such local maximum; the PSO described here, is designed to find the global maximum\cite{28}.
\subsection{Particle Swarm}
PSO searches through the fitness landscape by having particles known as agents fly, while attempting to find the optimal value. Each particle is randomly instantiated, and evaluated at each stage by the objective function. Thus, the particles represent candidate solutions. As previously stated, the PSO algorithm runs within a ``black - box'', and so all iterative refinements are based on the particles, with no awareness of the presence of abnormalities in the objective function\cite{25}.  

As shown above (figure 1), particles are created as candidate solutions on the fitness landscape. Each particle ($x_i$) is aware of its position ($r_i$), velocity ($v_i$), and the best position it has previously reached. Its position is comprised of both the candidate solution and its related fitness. The best position it previously reached is comprised of its individual best candidate and its individual best fitness. Similarly, the swarm as a single entity maintains its global best comprised of its global best candidate and its global best fitness\cite{25}. 

The PSO algorithm itself functions quite similarly to a parallel iterative refinement method. It repeats three steps until its stopping condition (e.g.  ${\mathcal E}\le \ .01$).
\begin{enumerate}
\item Evaluate current finesses
\item Update individual (and global) best positions
\item Update velocity and position for all agents
\end{enumerate}
Steps one and two have been covered above, with fitness being evaluated by evaluating the objective function with the candidate solution as an argument. Similarly, best fitness positions are updates by comparing the current best with the current evaluation for all individual particles and the swarm.  However, these steps are identical to those of a complete search of the entire fitness landscape. The third step gives particles a position and velocity based on the swarm's intelligence\cite{25}. 

The velocity update formula can be subdivided into three independent parts. First, the formula for the velocity at time $t+1$ must take the previous velocity (at time $t$) into account. However, this velocity is subject to change due to the particle's natural inertia. So, the first component of the formula is $wv_i(t)$, where w is a user supplied coefficient (generally $.8<w<1.2$). This inertial component forces the particle to continue travelling in the direction that it previously was. The exact value of $w$ allows the particle to experience acceleration or deceleration. Acceleration makes the swarm more likely to search a larger position of the fitness landscape, while deceleration usually makes the stopping condition be met sooner\cite{25}. 

The second term allows the particle to remember which areas of the fitness landscape have given more fruitful results. This cognitive component, $c_1r_1[\widehat{x_i}\left(t\right)-x_i\left(t\right)]$, is comprised of the cognitive coefficient $c_1$ (generally $c_1\approx 2$) and is related to the particle's affinity to move towards its individual best position ($\widehat{x_i}$). The third term, the social component takes the same form as the cognitive component, replacing the individual best position $\widehat{x_i}\ $with the global best position,$g(t)$, and the cognitive coefficient with a social coefficient ($c_2\approx c_1$). The resulting term, $c_2r_2[g(t)-x_i\left(t\right)]$,, is related to the particle's affinity to move towards the current global best position. Both terms represent the magnitude of the step that should be taken in a particular direction\cite{25}.  

The values $c_1$ and $c_2$ represent random coefficients in the cognitive and social components. These provide a stochastic nature to the velocity which makes particles move in a pseudo - random way that is constrained by the previous best solutions (individual and global)\cite{25}. 

A velocity clamp way be applied to prevent any particles from leaving the desired search space. For a search space bounded by $[-r_{max},\ r_{max}]$, the velocity clamp can be represented as $[-k\frac{r_{max}-r_{min}}{2},\ k\frac{r_{max}-r_{min}}{2}]$, where $k$ is a user - supplied parameter around\cite{8}.

The position of a particle is given in terms of its previous position and the velocity by:
\begin{center}
\begin{equation}
x_i\left(t+1\right)=x_i\left(t\right)+v_i\left(t+1\right)
\end{equation}
\begin{equation}
v_i\left(t+1\right)=wv_i\left(t\right)+c_1r_1\left[\widehat{x_i}\left(t\right)-x_i\left(t\right)\right]++c_2r_2[g(t)-x_i\left(t\right)]
\end{equation}
\end{center}
\subsection{Variations}
As previously stated, this process is repeated indefinitely till a stopping condition is met. Differences in the stopping condition has led to some variation among the various implementations of the PSO.  Other modifications include the inertial weight, which was not included in the velocity update, but has since become a standard component of the PSO. Other variations divide the swarm into separate populations, include an evolutionary ranking component, and stretching the objective function\cite{8}.

\section{Behaviour of Sparkling Squid}
\label{seq:behave}
\subsection{Behaviour of Sparkling Squid}
\textit{Watasenia scintillans}, or the sparkling squid is a small squid that lives around 1,200 feet deep in the Western Pacific Ocean.  The sparkling squid is one of the brightest of all the bioluminescent creatures of the sea. Every year during summer, Japan's shores are greeted to the amazing sight of the water glowing blue as millions of sparkling squid mate. Each squid has photophores across its body to attract other members of its species, and photophores on its tentacles and eyes to attract prey.  It is also able to use its photophores across its body for counter - illumination in an impressive form of camouflage.

The exact behavior and patterns of illumination of the sparkling squid remain unknown. In order to create a useful model that uses squid behavior, the ambiguity of the behavior of real squid has to be removed. To do so, we have divided the squid population into two sub - populations, a primary population and a secondary population. The primary population primarily moves on mutual attraction, while the secondary pursues unseen prey in a herd. The primary population operates on two basic assumptions:
\begin{enumerate}
 \item All squid are genderless; a less attractive squid will move toward a more attractive squid. The most attractive squid will move randomly.
\item The attractiveness of a squid is based on its brightness. The brightness of a squid is derived from its location on the fitness landscape.
\end{enumerate}
Similarly, the secondary population is modeled by two simple assumptions:
\begin{enumerate}
 \item The secondary population pursues unseen prey. The prey are more heavily populated in more fit areas of the fitness landscape.
\item Each individual squid seeks to improve its location by following the leader, and two random neighbors. Communication between members of the swarm may be imperfect.
\end{enumerate}

Using these characteristics, we have developed a stricter set of rules to model the sparkling squid population. We then use those rules to formulate an algorithmic procedure.

\subsection{Light Intensity}
We known that light intensity follows an inverse - square law with distance. Impurities in the water (and the water itself) cause the intensity of the light to decrease with distance. As stated previously, the brightness of the squid can be derived from the squid's location on the fitness landscape.

Under our model for the squid's behavior, only two factors have to be accounted for, the relation between attractiveness and intensity, and the change in light intensity as a function of distance. The second one of these two problems is much easier to solve as it requires a purely physical answer.  

We say that the distance between squid $i$ and $j$ is $r_{ij}$, and model the intensity in two parts. Firstly, we use an inverse - square law to represent the change in intensity with distance: $I(r) \propto \frac{1}{r_{ij}^2}$. The second part is the absorption of light in imperfect conditions, like the presence of water, impurities and murkiness. Assuming an absorption coefficient of $\gamma$, gives $I(r) \propto e^{-\gamma r_{ij}}$. These two models can be combined, assuming an original light intensity of $I_{0}$.
\begin{center}
\begin{equation}
I \left(r\right)=\frac{{I}_0}{1+\gamma r^2}
\end{equation}
\end{center}
This model also removes the singularity at $r_{ij}=0$. we chose to use a rational approximation over the Gaussian model to expedite computation. Finally, as the intensity of light decreases asymptotically, it is no longer noticeable by the squid. We use a piecewise function to make a fixed distinction between little light and no light.

\[I(r) = \left\{
  \begin{array}{lr}
    \frac{{I}_0}{1+\gamma r^2} & : x \le \ell \\
    0 & : x > \ell
  \end{array}
\right.
\]
The length scale $\ell$ is proportional to our choice of $\gamma$. Algorithmically, this separation speeds up the computation by ignoring extremely unfit solutions. While fine - tuning the length scale leads to better results, we use $\ell = \frac{1}{\gamma}$.
\section{Sparkling Squid Heuristics}
\label{sec:algos}
\subsection{Primary and Secondary Swarms}
The squid population is modeled in two parts, a primary population and a secondary population. The primary population is more fit, and therefore more attractive. These are the squid that are able to significantly influence the movement of other squid. The less - fit squid would be pushed and pulled around, without significantly influencing the movement of other squid. Therefore, we create a secondary squid population with those squid. The population division is dynamic, and is updated after every generation. 

We form the populations by ordering all squid $\textbf{x}_i\left(t\right)$ by $-f\left(\textbf{x}_i\right)$. Then, the primary population is $\{\textbf{x}_0\left(t\right),\cdots,\textbf{x}_{\frac{n}{2}}\left(t\right)\}$. Similarly, the secondary population is $\{\textbf{x}_{1+\frac{n}{2}}\left(t\right),\cdots,\textbf{x}_{n}\left(t\right)\}$.By dividing the populations in this way, we allow the primary swarm to function as a scout, explore the fitness landscape, while the secondary swarm provides leverage against local extrema, randomness and other abnormalities. It also reduces the runtime by a constant - factor of $\frac{1}{4}$. 

It is also worth noting that squid change population constantly. The least - fit squid in the primary swarm will most likely play the role of the leader in the secondary swarm, while the leader of the secondary swarm will be toyed with in the primary swarm. This communication allows the two separate swarms to function as a single unit. On a parallel machine, the population may be further subdivided to trade accuracy for speed.

\subsection{Attraction and Randomness}
We have chosen to make $\beta \left(\textbf{x}\right)\ \propto \ I(\textbf{x})$, and  $I \left(\textbf{x}\right)\ \propto \ f(\textbf{x})$. This means that for a given position vector $\textbf{x}$, $I_{0} \propto f(\textbf{x})$. By stacking all of our constants into  $\beta_{0}$ gives an attraction of:
\begin{center}
\begin{equation}
\beta \left(r,\textbf{x}\right)=\frac{{\beta_{0} f\left(\textbf{x}\right)}}{1+\gamma r^2}
\end{equation}
\end{center}
Algorithmically, we remove the position vector $\textbf{x}$ in favor of a pointer to the $i$-th squid. The attractiveness of squid $i$ is
$\beta_{0}$ gives an attraction of:
\begin{center}
\begin{equation}
\beta_{i} \left(r\right)=\frac{{\beta_{i_{0}}}}{1+\gamma r^2}
\end{equation}
\end{center}
Theoretically, $\gamma$ can hold any value that suits the test problem; we choose assign a standardized value to $\gamma$ that depends on $\overline{l}$, the average size of the search space in each dimension and $\overline{\beta_{0}}$, the average attractiveness at $r_{ij}=0$.
\begin{center}
\begin{equation}
\gamma =\frac{1}{g^2(1+{\ln  \left({\beta }_0\right)\ })}
\end{equation}
\end{center}

Each squid is modeled a random $d$-dimensional position vector. Based on the fitness of the function, the squid has a derived attractiveness from its position. The primary squid population then evolves without interference. However, the population is prone to finding local optima and getting stuck at nonoptimal locations. Therefore, we have included a random factor movement factor. The most attractive squid's movement are entirely governed by this factor. The random update term is based on a the simple idea of randomly moving the squid in either direction up to $.5$ units in every dimension. This is represented as:
\begin{center}
\begin{equation}
rand=\alpha \left(F(0,\ 1)-\frac{1}{2}\right),\ \alpha \in \left[0,\ 1\right]
\end{equation}
\end{center}
As solution quality improves, the random effects should decrease. This is accomplished by a randomness - reduction factor, $\delta$.
\begin{center}
\begin{equation}
random\left(t\right)=\alpha {\delta }^t\left(F(0,\ 1)-\frac{1}{2}\right),\ \alpha \in \left[0,\ 1\right],\delta \in (0,1]
\end{equation}
\end{center}
The lower the value of $\delta $, the faster the reduction, while a value of $\delta =1$ represents no reduction. The random value is generated according to the $F$, the arcsine distribution.

\subsection{Predator - Prey Swarms}
The secondary generation is far less individualized. Each squid looks toward the best squid, and two of its randomly chosen colleagues to choose a new position to move to. Its own position is only considered if the current squid is in the best position. Computationally, squid that are updated earlier are also considered in the update of later squid. At every stage, two distinct random squid (${x_r}_1$and ${x_r}_2$) are chosen. The next generation population is then created as:
\begin{center}
\begin{equation}
{{\mathbf x}}_{i}={{\mathbf x}}_{best(t)}+F({{\mathbf x}}_{r1}-{{\mathbf x}}_{r2})
\end{equation}
\end{center}
$F$, is a mutation factor in the range $F\in [0,\ 2]$. We can also add a requirement for the new position of the squid to be better than the previous one for it to move. It is worth noting by centering the new swarm around the previous best, a non - zero number of squid are expected to switch swarms. This helps lagging squid in the primary swarm escape local optima, and squid in the secondary swarm explore more parts of the fitness landscape.
\subsection{Sparkling Squid Algorithms}
Based on our assumptions and model, we can now outline the basic sparkling squid algorithm. Note that in implementation, distances ($r_{ij}$) can be calculated using any metric in $d$ - dimensions.  We choose to use the $\ell^{2}$ Norm, or Euclidean Norm. It might also be interesting to use other norms, such as the Frobenius Norm, or Hilbert - Schmidt Norm in certain applications.
\begin{algorithm}[htbp]
\caption{Approximate $X_G$}
\begin{algorithmic}
\REQUIRE Initial solution to a $d$ - dimensional problem: $X_i$
\STATE Create initial squid population: $X$
\STATE Evaluate fitness of squid population: $f(X)$
\STATE $X_G \leftarrow max(f(X))$
\FOR{$t \leftarrow 1$ to $max$}
\STATE Sort $X$ by $-f(X)$
\STATE $S_1 \leftarrow \left\{X_0, \dots, X_\frac{n}{2}\right\}$
\STATE $S_2 \leftarrow \left\{X_{1+\frac{n}{2}}, \dots, X_n\right\}$
\FOR{$i \leftarrow 0$ to $\frac{n}{2}$}
\FOR{$j \leftarrow 0$ to $\frac{n}{2}$}
\IF{$f(X_i)>f(X_j)$}
\STATE Calculate distance and attraction
\STATE Update position and position - based values
\ENDIF
\ENDFOR
\ENDFOR
\FOR{$i \leftarrow 1+\frac{n}{2}$ to $n$}
\STATE Update position $X_{i}$
\ENDFOR
\STATE $X_G \leftarrow$ Current best solution
\STATE $t \leftarrow t+1$
\ENDFOR
\end{algorithmic}
\end{algorithm}
\section{Multimodal Optimization}
\label{sec:multi}
\subsection{Benchmark Validation}
We have used a set of $3$ standard functions for the purpose of validation \cite{50,29,59,89}. Beale's function, Easom's fucntion and Michalewicz's function are all standard functions for the purpose of validating and comparing new algorithms. The Beale function
\begin{center}
\begin{equation}
f(\textbf{x})=(1.5-x_1+x_1x_2)^2+(2.25-x_3+x_1x_2^2)^2+(2.625-x_1+x_1x_2^3)^2, -4.5 \le x_i \le 4.5
\end{equation}
\end{center}
has $f(\textbf{x}^*)=0$ at $\textbf{x}^*=(3,0.5)$. 

\includegraphics[width=\linewidth]{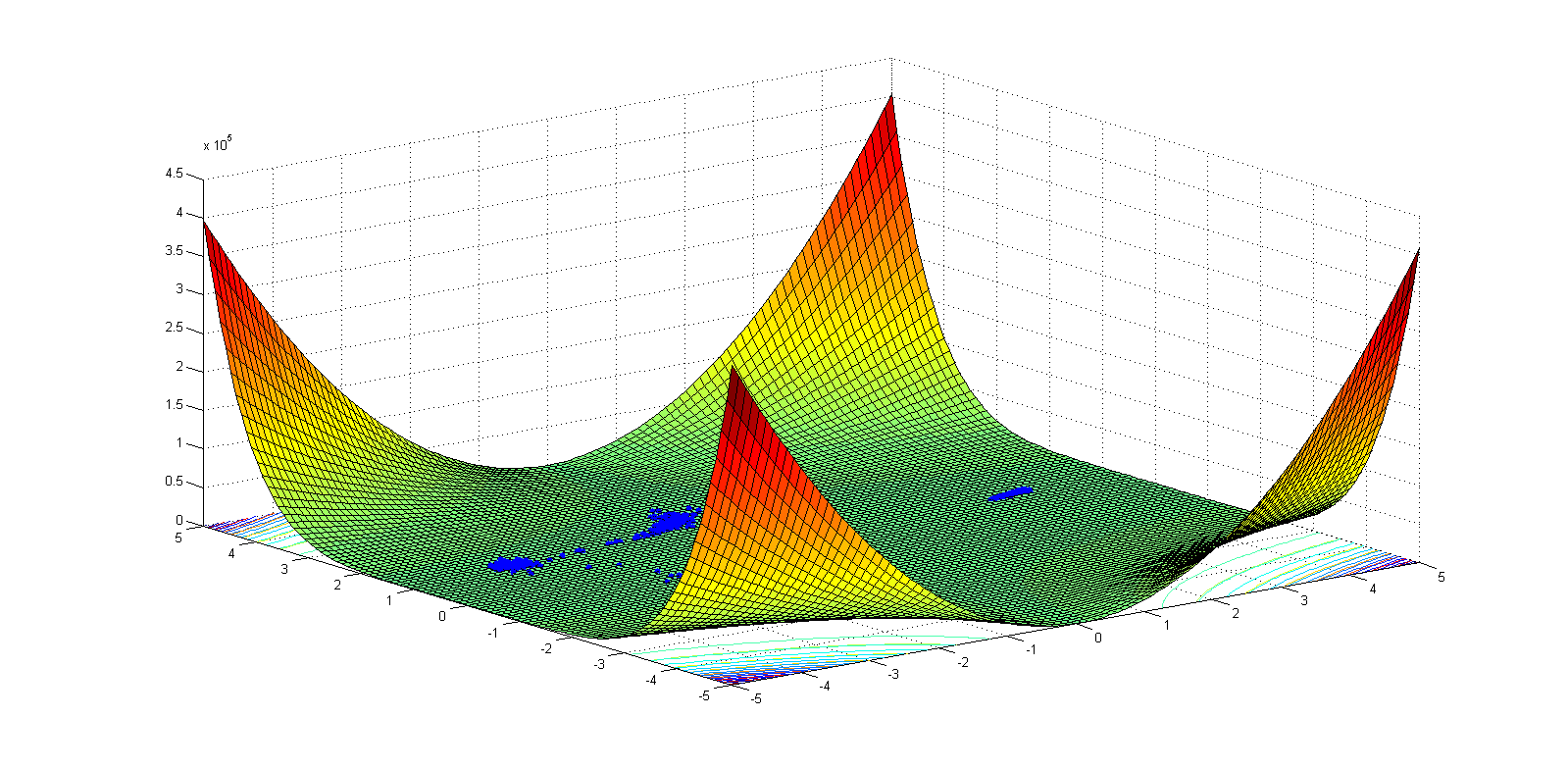}
\hspace*{15pt}\hbox{\scriptsize Credit:\thinspace{\small\itshape Videh Seksaria}}
  \captionof{figure}{Beale's Function, with squid paths in blue}

Easom's function
\begin{center}
\begin{equation}
g(\textbf{x})=-\cos(x_1)\cos(x_2)\exp(-(x_1-\pi)^2-(x_2-\pi)^2), -100 \le x_i \le 1000
\end{equation}
\end{center}
has $g(\textbf{x}^*)=-1$ at $\textbf{x}^*=(\pi,\pi)$.

\includegraphics[width=\linewidth]{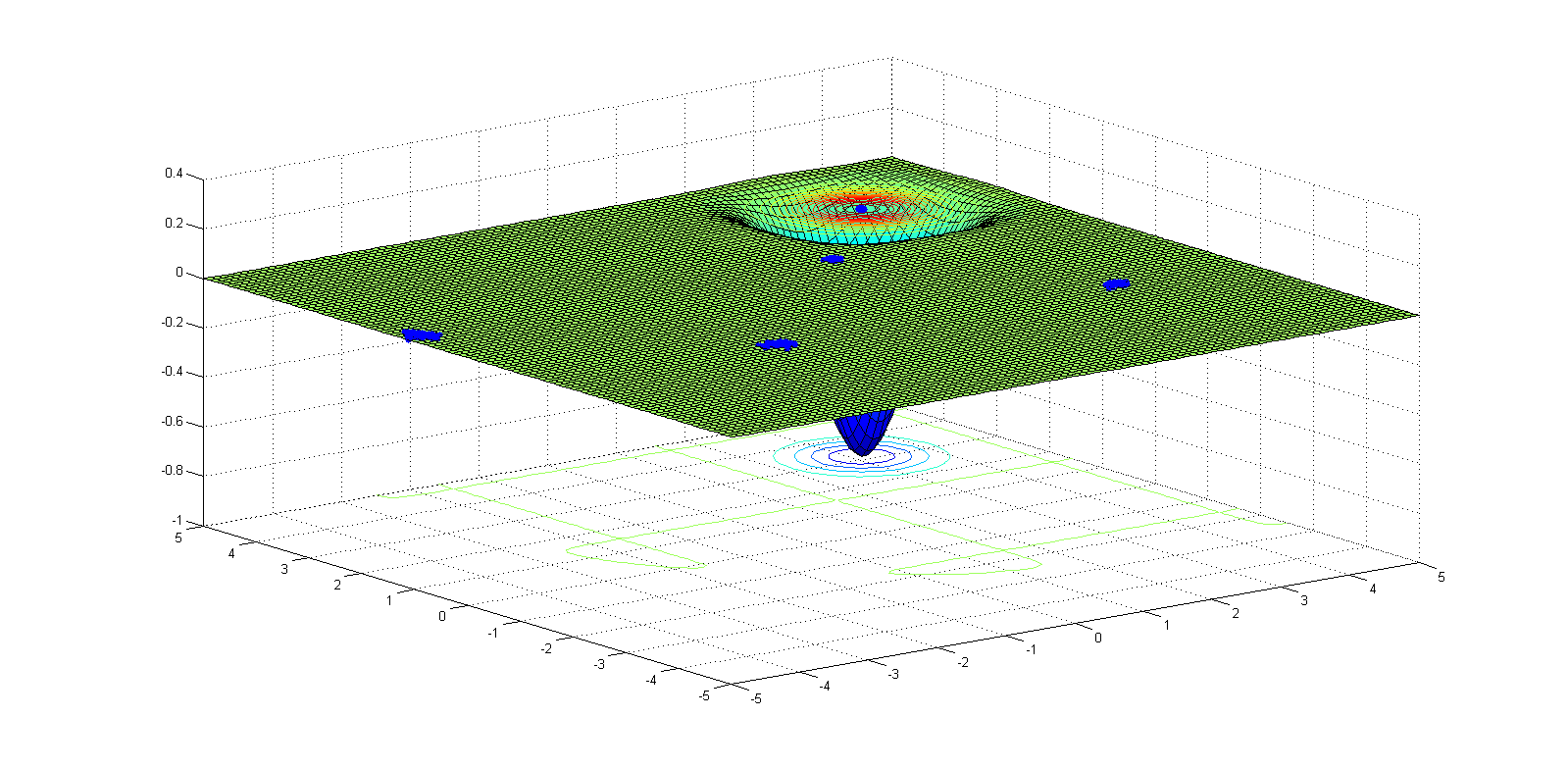}
\hspace*{15pt}\hbox{\scriptsize Credit:\thinspace{\small\itshape Videh Seksaria}}
  \captionof{figure}{Easom's Function, with squid paths in blue}
  
 Finally, Michalewicz's function
\begin{center}
\begin{equation}
-\sum\limits_{i=1}^d \sin(x_i)\sin^{2m}\left(\frac{ix_i^2}{\pi}\right), m = 10, 0 \le x_i \le \pi
\end{equation}
\end{center}
for $n=10$ has $h(\textbf{x}^*)=-9.66015$. In all three cases, the SSA found the global minimum to within $\epsilon \le 1*10^-5$ within 90 seconds. This was true in over $10,000$ trials. 

In the case of the Beale function, the SSA was able to escape the four sharp peaks around the corners within (approx.) $500$ evaluations, and the the valley ridges within (approx.) $1,750$ evaluations. When tested with the Easom function, the SSA spent (approx.) $750$ evaluations in various ridges, before finding the global minima in (approx.) $2,000$ evaluations. Finally, on the Michaelwicz function, the SSA spent (approx.) $1,000$ evaluations going between the various local minima, and hill climbing. Finally, after (approx.) $3,500$ evaluations it found the global minima.

Across all $3$ benchmark functions, the SSA was able to find the global minima in time comparable to the PSA and GA. Given additional evaluations, the solution quality increases negligibly. 
\subsection{Comparative Results with Particle Swarm and Genetic Algorithms}
We have chosen a set of eight common benchmark function to evaluate the performance of SSA against other common evolutionary algorithms\cite{50,29,59,89}. Due to the existence of countless modifications and improvements to the particle swarm and genetic algorithms, we have implemented them in their original forms. For PSO, we use the standard implementation of Kennedy and Eberhart, with $\alpha=2$, and a constant inertial term of $I=1$.For GA, we ignore the effects of elitism, use a mutation probability $p_m=0.05$ and a crossover probability of $0.95$\cite{18,23,51,4,31,52}.

\par There are many valid ways of running such a comparision test; we have opted to run each algorithm $10,000$ times per function and report and best, worst, mean and standard deviation of the global minima reported. In the interest of fairness, each algorithm was given $1.50$ seconds to run. In addition, comparable population sizes, etc. were used across all three algorithms. Previous analysis of the SSA, PSO and GA were used to define optimal parameters under the testing conditions\cite{20,30,17,16,11}.

Simulations were completed using MATLAB, with the code for PSO and GA taken directly from their source paper. All tests were completed on an Intel i7-2670QM with 2.2GHz per each of 4 cores. Tests were completed with constant minimal background and system processes. The data is presented in tables following a brief description of the trial function, followed by results.

The objective function for the first test problem is the Ackley Function. This problem is given as:
\begin{center}
\begin{equation}
f\left(\textbf{x}\right) = -20\exp\left(-0.2 \sqrt{\frac{1}{n} \sum_{i=1}^n x_i^2}\right) - \exp\left(\frac{1}{n} \sum_{i=1}^n cos\left(2\pi x_i\right)\right) + 20 + e
\end{equation}
\end{center}
It has $f(\textbf{x}^*)=0$ at $\textbf{x}^*=(0,\cdots,0)$. 
\begin{table}[!htbp] 
\caption{Comparison among PSO, GA and SSA on Ackley's Function($d=128$)}  
\centering
\begin{tabular}{c c c c c c c} 
\hline
\hline
Metric &        & PSO &        & GA &        & SSA \\ [0.5ex]   
\hline
Best &        & 8.99e-08 &        & 6.76e-07 &        & 4.52e-08 \\ 
Worst &        & 7.99e-06 &        & 8.17e-07 &        & 1.62e-07 \\ 
Mean &        & 3.31e-07 &        & 4.66e-06 &        & 9.10e-08 \\ 
Std. Dev.&        & 5.81e-07 &        & 2.41e-06 &        & 4.31e-08 \\ 
Rank &        & 2 & & 3 &        & 1 \\ [1ex] 
\hline 
\end{tabular} 
\label{table:ackley} 
\end{table} 
All three algorithms were able to circumvent the numerous local minima in Ackley's function, and find the valley containing the global minima. The reported minima from each algorithm depends on how quickly the algorithms could run. There is a negligible difference between the PSO and SSA, with the GA as a close third.

For the second test problem, we used the Goldstein - Price Function. It is a function of two - variables, with several local minima and a large fitness landscape. It is defined as
\begin{center}
\begin{equation}
\begin{split}
f\left(\textbf{x}\right) = \left(1+\left(x_1+x_2+1\right)^2\left(19-14x_2+3x_1^2-14x_2+6x_1x_2+3x_2^2\right)\right) \\ \left(\left(30+2x_1-3x_2\right)^2\left(18-32x_1+12x_1^2+48x_2-36x_1x_2+27x_2^2\right)\right)-3
\end{split}
\end{equation}
\end{center}
It has $f(\textbf{x}^*)=0$ at $\textbf{x}^*=(0,1)$. 

\begin{table}[!htbp] 
\caption{Comparison among PSO, GA and SSA on Goldstein and Price's Function}  
\centering
\begin{tabular}{c c c c c c c} 
\hline
\hline
Metric &        & PSO &        & GA &        & SSA \\ [0.5ex]   
\hline
Best &        & 6.21e-08 &        & 2.89e-07 &        & 2.71e-08 \\ 
Worst &        & 8.92e-07 &        & 4.09e-06 &        & 8.12e-08 \\ 
Mean &        & 5.38e-07 &        & 2.52e-06 &        & 5.12e-08 \\ 
Std. Dev.&        & 3.07e-07 &        & 1.31e-06 &        & 1.62e-8 \\ 
Rank &        & 2 & & 3 &        & 1 \\ [1ex] 
\hline 
\end{tabular} 
\label{table:goldprice} 
\end{table}

\includegraphics[width=\linewidth]{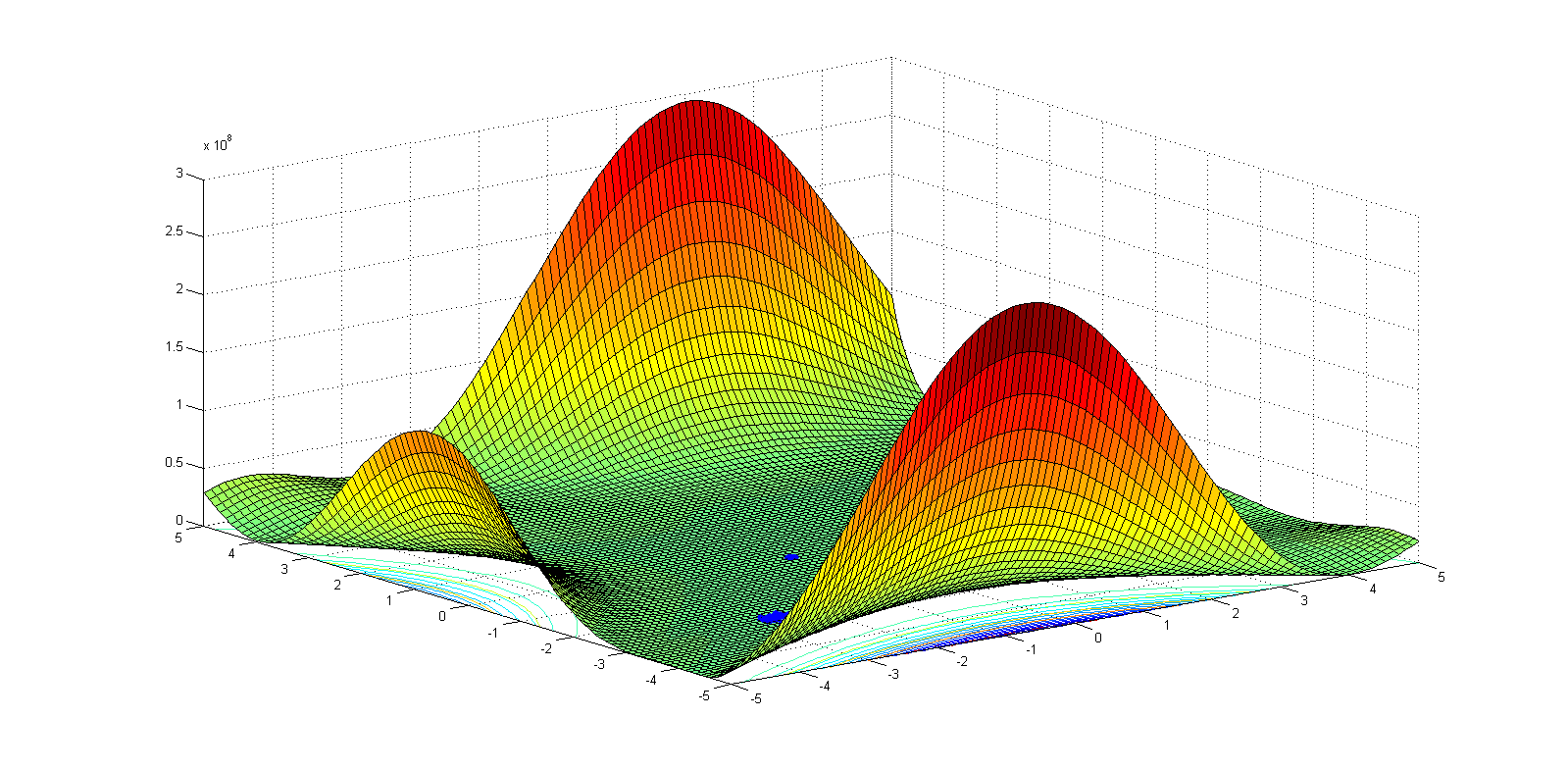}
\hspace*{15pt}\hbox{\scriptsize Credit:\thinspace{\small\itshape Videh Seksaria}}
  \captionof{figure}{Beale's Function, with squid paths in blue}

Like with Ackley's Function, the SSA and PSO came in first and second place respectively, with the GA in third. This was expected due to the similarity between the two functions. As above, all three algorithms located the valley and correct ridge with the global minima. The output value depended on the efficiency of the algorithm,

The objective function for the third test problem is the Griewank Function. This problem is given as:
\begin{center}
\begin{equation}
f\left(\textbf{x}\right) = 1 + \frac{1}{4000} \sum_{i=1}^n x^2_i - \prod_{i=1}^n \cos\left(\frac{x_i}{\sqrt{i}}\right)
\end{equation}
\end{center}
It has $f(\textbf{x}^*)=0$ at $\textbf{x}^*=(0,\cdots,0)$. 

\begin{table}[!htbp] 
\caption{Comparison among PSO, GA and SSA on Griewank's Function}  
\centering
\begin{tabular}{c c c c c c c} 
\hline
\hline
Metric &        & PSO &        & GA &        & SSA \\ [0.5ex]   
\hline
Best &        & 5.92E-07 &        & 1.51E-08 &        & 3.59E-08 \\ 
Worst &        & 7.99E-06 &        & 8.85E-07 &        & 9.10E-08 \\ 
Mean &        & 2.56E-06 &        & 3.20E-07 &        & 6.18E-08 \\ 
Std. Dev.&        & 2.98E-07 &        & 1.31e-06 &        & 1.93E-08 \\ 
Rank &        & 3 & & 2 &        & 1 \\ [1ex] 
\hline 
\end{tabular} 
\label{table:griewank} 
\end{table}
Once again, all three algorithms were able to find the valley with the global minima, but PSO got stuck on local ridges in most trials. The GA and SSA algorithms performed very similarly, due to their ability to escape such footholds and find the global minima. The difference between the GA and SSA in minuscule, while the PSO lags behind in third.

The objective function for the third test problem is the Levy Function. This problem is given as:
\begin{center}
\begin{equation}
\begin{split}
f\left(\textbf{x}\right) = \sin^2\left(\frac{\pi x_1+3\pi}{4}\right)+ \\ \sum\limits_{i=1}^{d-1}{\left(\frac{x_1-1}{4}\right)^2\left(1+10\sin^2\left(\frac{\pi x_1+7\pi}{4}\right)\right)+\left(\frac{x_d-1}{4}\right)^2\left(1+sin^2\left(\frac{2\pi x_d+6\pi}{4}\right)\right)}
\end{split}
\end{equation}
\end{center}
It has $f(\textbf{x}^*)=0$ at $\textbf{x}^*=(1,\cdots,1)$. 
\begin{table}[!htbp] 
\caption{Comparison among PSO, GA and SSA on Levy's Function(d=16)}  
\centering
\begin{tabular}{c c c c c c c} 
\hline
\hline
Metric &        & PSO &        & GA &        & SSA \\ [0.5ex]   
\hline
Best &        & 3.06E-08 &        & 2.16E-08 &        & 3.03E-08 \\ 
Worst &        & 9.52E-08 &        & 9.76E-07 &        & 9.51E-08 \\ 
Mean &        & 6.56E-08 &        & 5.48E-07 &        & 5.83E-08 \\ 
Std. Dev.&        & 2.61E-08 &        & 3.67E-07 &        & 1.93E-08 \\ 
Rank &        & 2 & & 3 &        & 1 \\ [1ex] 
\hline 
\end{tabular} 
\label{table:levy} 
\end{table}
Levy's Function had some of the closest results in all of out tests. The jagged landscape of the Levy Funciton proved to be no challenge for any of our three algorithms. All were able to find the foxhole with the global minima and reached quite close to the numerical minima.

For the fith test, we chose the popular Rastrigin Function. This problem is given as:
\begin{center}
\begin{equation}
f\left(\textbf{x}\right) = 10n + \sum_{i=1}^n \left(x_i^2 -10\cos\left(2\pi x_i\right)\right)
\end{equation}
\end{center}
It has $f(\textbf{x}^*)=0$ at $\textbf{x}^*=(0,\cdots,0)$.
\begin{table}[!htbp] 
\caption{Comparison among PSO, GA and SSA on Rastrigin's Function)}  
\centering
\begin{tabular}{c c c c c c c} 
\hline
\hline
Metric &        & PSO &        & GA &        & SSA \\ [0.5ex]   
\hline
Best &        & 4.78E-13 &        & 1.57E-07 &        & 2.18E-08 \\ 
Worst &        & 1.04E-08 &        & 3.86E-06 &        & 1.39E-07 \\ 
Mean &        & 1.68E-09 &        & 1.01E-06 &        & 8.59E-08 \\ 
Std. Dev.&        & 3.12E-09 &        & 1.43E-06 &        & 3.67E-08 \\ 
Rank &        & 1 & & 3 &        & 2 \\ [1ex] 
\hline 
\end{tabular} 
\label{table:rastrigin} 
\end{table}

\includegraphics[width=\linewidth]{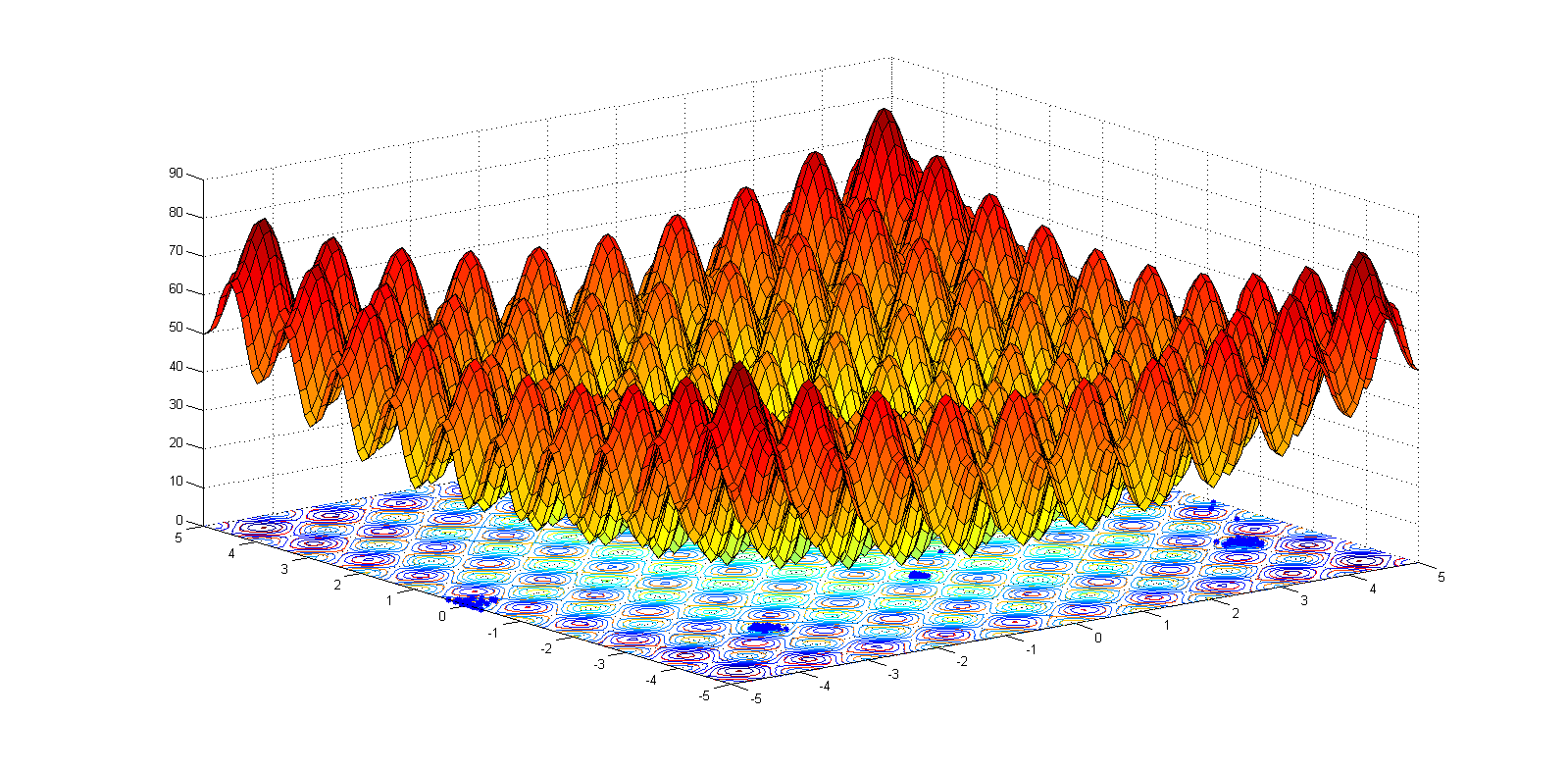}
\hspace*{15pt}\hbox{\scriptsize Credit:\thinspace{\small\itshape Videh Seksaria}}
  \captionof{figure}{Beale's Function, with squid paths in blue}
  
The Rastrigin Function is one of two test cases where the PSO shined. Here, PSO came in a clear first place followed by the SSA and GA in a distant second and third place. The PSO was able to find the ridge with the global minima consistently, while the SSA spent most of its time in near - by local minima. Similarly, the GA went from local minima to local minima, never even reaching the correct ridge.

For the sixth test, we chose the popular Rosenbrock Function. Aptly known as Rosenbrock's Valley, this test function features a narrow, parabolic valley. This problem is given by:
\begin{center}
\begin{equation}
f\left(\textbf{x}\right) = \sum_{i=1}^{n-1} \left(100\left(x_i^2 - x_{i+1}\right)^2 + \left(1-x_i\right)^2\right)
\end{equation}
\end{center}
It has $f(\textbf{x}^*)=0$ at $\textbf{x}^*=(1,\cdots,1)$
\begin{table}[!htbp] 
\caption{Comparison among PSO, GA and SSA on Rosenbrock's Function(d=128)}  
\centering
\begin{tabular}{c c c c c c c} 
\hline
\hline
Metric &        & PSO &        & GA &        & SSA \\ [0.5ex]   
\hline
Best &        & 7.94E-07 &        & 3.00E-08 &        & 4.66E-08 \\ 
Worst &        & 1.06E-05 &        & 9.00E-08 &        & 1.60E-07 \\ 
Mean &        & 3.72E-06 &        & 8.61E-08 &        & 7.77E-08 \\ 
Std. Dev.&        & 3.49E-06 &        & 2.45E-08 &        & 3.25E-08 \\ 
Rank &        & 3 & & 2 &        & 1 \\ [1ex] 
\hline 
\end{tabular} 
\label{table:rosenbrock} 
\end{table}
The Rosenbrock Function was our only test case where the GA did not come in third place. The SSA came in first by an extremely small margin, while the GA defeated the PSO by a large margin. All three algorithms were able to find the Rosenbrock valley, but the PSO frequently got stuck in an nonoptimal area of the valley. The results of the GA and SSA, with the SSA coming ahead by exploring a larger amount of the valley. This also resulted a worse worst output compared to the GA.

For the seventh test, we chose the Shekel Function. The Shekel Function, is often known as the foxhole function because of its minima appearing like foxholes on the fitness landscape. It is defined in terms of a vector $\beta$ and a matrix $C$.
\begin{center}
\begin{equation}
f\left(\textbf{x}\right) =-\sum\limits_{i=1}^d {\left(\sum\limits_{j=1}^4{\left(x_j-C_{ji}\right)^2}+\beta_i\right)}^{-1}
\end{equation}
\end{center}
\begin{center}
$\beta=\frac{1}{10}(1,2,2,4,4,6,3,7,5,5)^{T}$
$\textbf{C}=\left( \begin{array}{cccccccccc}4.0&1.0&8.0&6.0&3.0&2.0&5.0&8.0&6.0&7.0\\4.0&1.0&8.0&6.0&7.0&9.0&3.0&1.0&2.0&3.0\\4.0&1.0&8.0&6.0&3.0&2.0&5.0&8.0&6.0&7.0\\4.0&1.0&8.0&6.0&7.0&9.0&3.0&1.0&2.0&3.0\end{array} \right)$
\end{center}
It has $f(\textbf{x}^*)=0$ at $\textbf{x}^*=(4,4,4,4)$. We normalized the Shekel function so that the global minimum is $f(\textbf{x}^*)=0$, instead of $f(\textbf{x}^*)\approx -10.5364$
\begin{table}[!htbp] 
\caption{Comparison among PSO, GA and SSA on Shekel's (Foxhole) Function(d=10)}  
\centering
\begin{tabular}{c c c c c c c} 
\hline
\hline
Metric &        & PSO &        & GA &        & SSA \\ [0.5ex]   
\hline
Best &        & 2.00E-07 &        & 3.52E-08 &        & 5.66E-08 \\ 
Worst &        & 2.00E-07&        & 8.63E-07 &        & 2.81E-07 \\ 
Mean &        & 2.00E-07 &        & 2.22E-07 &        & 1.38E-07 \\ 
Std. Dev.&        & 0.00E+0 &        & 2.43E-07 &        & 1.01E-07 \\ 
Rank &        & 2 & & 3 &        & 1 \\ [1ex] 
\hline 
\end{tabular} 
\label{table:shekel} 
\end{table}
The results for the Shekel Funciton are especially interesting. The PSO show no variation in all $10,000$ trails runs. This is due to it always taking an identical path around the idential set of foxholes, and finally reaching (and stopping) in a particular ridge near the global minima. Even so, the PSO came in second place, compared to the SSA in first and GA in third. All three algorithms located the correct foxhole, and their precision depended on efficiency.

Finally, we also used DeJong's First Function. Usually called the Sphere Function, it is continuous, convex, unimodal $d-$dimensional hypersphere, with $f(\textbf{x}^*)=0$ at $\textbf{x}^*=(0, \cdots, 0)$.
\begin{center}
\begin{equation}
f\left(\textbf{x}\right) =\sum_{i=1}^n x_i^2
\end{equation}
\end{center}
It has $f(\textbf{x}^*)=0$ at $\textbf{x}^*=(0,\cdots,0)$
\begin{table}[!htbp] 
\caption{Comparison among PSO, GA and SSA on De Jong's First Function (Sphere)(d=32)}  
\centering
\begin{tabular}{c c c c c c c} 
\hline
\hline
Metric &        & PSO &        & GA &        & SSA \\ [0.5ex]   
\hline
Best &        & 4.08E-14 &        & 2.89E-08 &        & 3.49E-08 \\ 
Worst &        & 3.85E-10 &        & 8.57E-07 &        & 1.90E-07 \\ 
Mean &        & 4.02E-11 &        & 1.60E-07 &        & 6.65E-08 \\ 
Std. Dev.&        & 1.15E-10 &        & 2.47E-07 &        & 4.53E-08 \\ 
Rank &        & 1 & & 3 &        & 2 \\ [1ex] 
\hline 
\end{tabular} 
\label{table:dejong} 
\end{table}
On the Sphere Function, all three algorithms quickly located the global minima. Even at worst, they were able to report a precision of $1E-7$. The PSO reported an amazingly accurate $1E-11$ on average in the allotted time, with the SSA and GA close behind. The Sphere Function showed the PSO's power in minimizing simple functions like the Sphere Function.

Finally, we have reported a summary of the results of the above tables. In calculating the success rate, we used a tolerance of $5E-7$, assuming a normal distribution on the mean and standard deviation.
\begin{table}[!htbp] 
\caption{Comparison among PSO, GA and SSA}  
\centering
\begin{tabular}{c c c c c c c} 
\hline
\hline
&        & PSO &        & GA &        & SSA \\ [0.5ex]   
\hline
Average &        & 2.00 &        & 2.75 &        & 1.25 \\ 
Final &        & 2 &        & 3 &        & 1 \\ 
Success Rate &        & 93.88\% &        & 94.05\% &        & 100\% \\  
\hline 
\end{tabular} 
\label{table:final} 
\end{table}
After all $8$ tests, 
\section{Conclusions}
\label{sec:conc}
We have taken the model of the behavior of the sparkling squid and successfully formulated a new Sparkling Squid Algorithm (SSA) from it. From initial testing and benchmark evaluations, we have seen that it is a powerful and promising algorithm. More specifically, we found that the SSA is able to outperform both PSO and GA at optimizing various functions. This makes it probable that the SSA will be useful in solving NP - hard problems, a topic for further investigation. In addition, looking at the convergence behaviour of the SSA shows that it is able to escape from local optima quite well.

Although quite efficient the SSA is not specifically tailored to any problem. One possible modification would be the addition of an adaptive control parameter. Secondly, various methods of providing direction to the squid could improve overall performance. More simply, fine - tuning of the various parameters ($\alpha$, etc.) can improve overall performance. It might also be interesting to apply the SSA to various NP - Hard problems, such as the Traveling Salesman Problem. Lastly, the combination of the SSA with existing optimization paradigms such as the PSO and GA may prove fruitful.

\end{document}